\pdfoutput=1

\documentclass[11pt]{article}

\usepackage[preprint]{acl}
\usepackage{times}
\usepackage{latexsym}
\usepackage[T1]{fontenc}
\usepackage[utf8]{inputenc}
\usepackage{microtype}
\usepackage{inconsolata}

\usepackage{soul}
\usepackage{url}
\usepackage{graphicx}
\usepackage{amsmath}
\usepackage{amsthm}
\usepackage{booktabs}
\usepackage{algorithm}
\usepackage{algorithmic}
\usepackage{multirow}
\usepackage{amssymb}
\usepackage{subcaption}
\usepackage{colortbl}
\usepackage{paralist}
\usepackage{makecell}
\usepackage{arydshln}

\title{\textit{Ten Words Only Still Help:} Improving Black-Box AI-Generated Text Detection via Proxy-Guided Efficient Re-Sampling}

\newcommand{\emailsI}{\href{mailto:shiyuhui22s@ict.ac.cn}{shiyuhui22s},\href{mailto:shengqiang18z@ict.ac.cn}{shengqiang18z},\href{mailto:caojuan@ict.ac.cn}{caojuan},\href{mailto:hubeizhe21s@ict.ac.cn}{hubeizhe21s},\href{mailto:wangdanding@ict.ac.cn}{wangdanding}}

\author{
Yuhui Shi$^{1,2}$\quad
Qiang Sheng$^{1}$\quad
Juan Cao$^{1,2}$\quad
Hao Mi$^{1,3}$\quad
Beizhe Hu$^{1,2}$\quad
Danding Wang$^{1}$\\
$^{1}$Key Lab of Intelligent Information Processing of Chinese Academy of Sciences, \\
	Institute of Computing Technology, Chinese Academy of Sciences\\
$^{2}$University of Chinese Academy of Sciences \quad 
$^{3}$Xi'an Jiaotong University \\
\texttt{\{\emailsI\}@ict.ac.cn}\\
\texttt{\href{mailto:mihao1018@gmail.com}{mihao1018}@gmail.com}
}

\begin{document}
\maketitle
\begin{abstract}
With the rapidly increasing application of large language models (LLMs), their abuse has caused many undesirable societal problems such as fake news, academic dishonesty, and information pollution. This makes AI-generated text (AIGT) detection of great importance. Among existing methods, white-box methods are generally superior to black-box methods in terms of performance and generalizability, but they require access to LLMs' internal states and are not applicable to black-box settings. In this paper, we propose to estimate word generation probabilities as pseudo white-box features via multiple re-sampling to help improve AIGT detection under the black-box setting. Specifically, we design \textbf{POGER}, a proxy-guided efficient re-sampling method, which selects a small subset of representative words (e.g., 10 words) for performing multiple re-sampling in black-box AIGT detection. Experiments on datasets containing texts from humans and seven LLMs show that POGER outperforms all baselines in macro F1 under black-box, partial white-box, and out-of-distribution settings and maintains lower re-sampling costs than its existing counterparts. 
\end{abstract}

\section{Introduction}
\begin{figure}[t]
\centering
    \includegraphics[width=\linewidth,trim=100 90 120 15,clip]{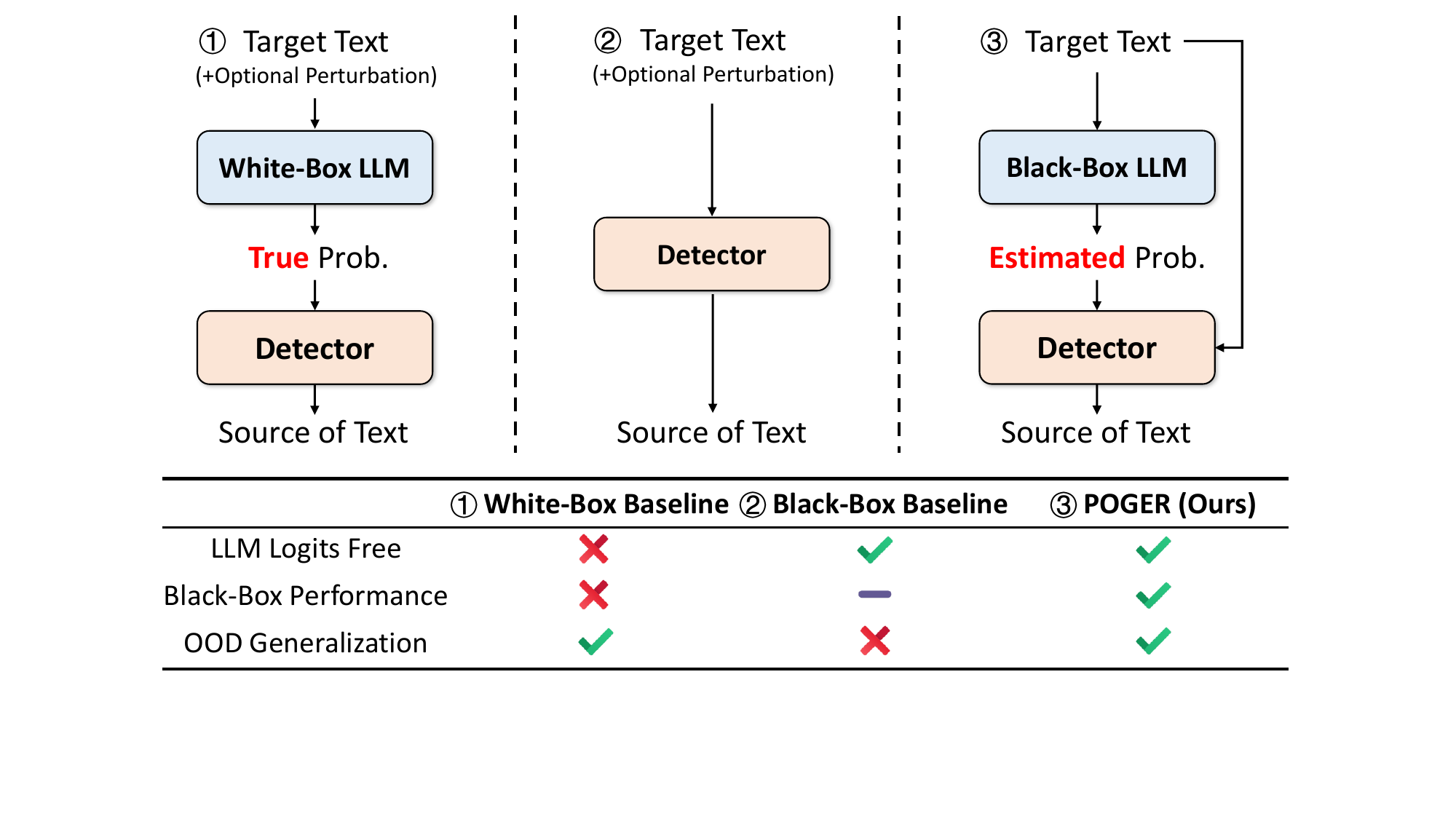}
    \caption{Paradigm comparison between our proposed POGER and existing white-box/black-box methods. POGER does not require LLMs' internal states like output logits and performs better than the other types of baselines under black-box and out-of-distribution (OOD) settings.}
    \label{fig:result}
\end{figure}

Recent breakthroughs in large language models (LLMs) have significantly improved the quality of AI-generated text (AIGT) and further boosted applications in diverse scenarios. People can easily instruct LLM-supported services like ChatGPT~\cite{chatgpt} to generate texts that are almost imperceptible to humans~\cite{jakesch2023human,uchendu2023does}. Though LLMs bring much convenience, new societal threats caused by their abuses also emerged: Political manipulators produce AI-generated fake news to risk democracy~\cite{lucas2023fighting}. Students cheat by submitting AI-generated works without paying expected efforts~\cite{bohacek2023unseen}. And content farms accelerate information pollution with AI-generated low-quality articles~\cite{Brewster_Fishman_Xu_2023}. To build the first barrier against such threats, developing techniques for detecting AI-generated text is of urgent need.

AI-generated text detection is generally defined as a binary or multiclass classification task. The former is to distinguish human-written and AI-generated text, while the latter subsequently recognizes which LLM generates the given text (usually for forensics needs). According to whether the detector can access the source LLM's internal states~\cite{yang2023survey}, existing methods can be categorized as white-box and black-box methods.
White-box methods~\cite{detectgpt,sniffer} distinguish LLMs using delicate features reflected by internal states like output token probabilities and usually achieve high detection performance, but its applicability is limited due to the widely existing unavailability of internal states in commercial LLM services.
In contrast, black-box methods~\cite{hc3} require output texts only for feature acquirement and in principle could be applied to any LLM. However, they generally underperform white-box methods and are more likely to suffer generalization issues on texts from a new domain~\cite{conda}. Such a dilemma poses a key challenge for effectively detecting AI-generated text in reality.

\textbf{To address this issue, a possible solution is to estimate features that proved effective in white-box detection for black-box scenarios} (see \figurename~\ref{fig:result}).
Inspired by the recent study on inferring the decoding strategy of an LLM with the multiple re-sampling~\cite{ippolito2023reverse}, we suppose that the statistics on re-sampling results on black-box LLMs could serve as a good estimation of word output probabilities, which reflects the nuance of internal states among different LLMs and subsequently help improve black-box detection.
In this paper, we conduct the first empirical exploration along this line.
Preliminarily, we implement a naive but costly solution that prompts the LLM with a continuation instruction multiple times on each position of the given text (\emph{i.e.}, full-text re-sampling). Results show that even using \textit{estimated} word probabilities, the detector still outperforms the typical black-box RoBERTa-based detector by 14.3\%, validating the feasibility of re-sampling-based black-box detection.

To reduce the sampling cost and improve the practicality, we further design \textbf{POGER}, a proxy-guided efficient re-sampling method for black-box AIGT detection.
The core idea of POGER is to select a subset of words possibly indicative of the LLMs' unique word use characteristics from the given text.
As LLMs are usually trained on large-scale human language corpora, the word generated with high probability is often similar across different LLMs under the same context, reflecting human language preferences. Instead, words with lower probabilities are more likely to expose LLMs' uniqueness.
Therefore, we employ a proxy white-box LLM to nominate words of relatively low probabilities across the text sequence and then preserve the words with low probability estimation error.
By performing re-sampling only for the positions of these words, POGER largely reduces the required sampling times while still maintaining the advantages over existing methods in the challenging 8-class black-box setting. Our contributions are as follows:

\begin{compactitem}
    \item We propose to use estimated word generation probabilities to empower black-box AI-generated text detection and empirically show its feasibility.
    \item We design POGER, a proxy-guided efficient re-sampling method that 
    largely reduces sampling cost and maintains detection performance by recognizing words that reflect LLMs' uniqueness.
    \item Extensive experiments on texts from humans and 7 popular LLMs show the superiority of POGER over existing methods for binary, multiclass, and out-of-distribution detection scenarios.\footnote{\url{https://github.com/ICTMCG/POGER}}
\end{compactitem}

\section{Background}
\subsection{Task Formulation}
Given a text containing $n$ words $\boldsymbol{x}=(x_1,x_2,\cdots,x_n)$, AIGT detection aims to obtain a classifier $f:\boldsymbol{x}\rightarrow y$, where $y$ is the source of $\boldsymbol{x}$.
The task can be further categorized into:

\textbf{1) Binary AIGT Detection:} Distinguish whether a text is generated by AI, \emph{i.e.}, $y\in\left\{\textrm{human}, \textrm{AI}\right\}$.

\textbf{2) Multiclass AIGT Detection:} Distinguish where a text is from human or a specific AI model, \emph{i.e.}, $y\in\left\{\textrm{human}, \theta_1, \theta_2, \cdots, \theta_M\right\}$, where $\theta_i$ is an AI model that can generate text. Similar concepts include origin tracing~\cite{sniffer} and authorship attribution~\cite{uchendu2020authorship}.

\subsection{Related Works}
The detection of AIGT can be active or passive. Active methods add pre-designed watermarks to LLM-generated text for later identification~\cite{pmlr-v202-kirchenbauer23a,yoo-etal-2023-robust,liu2023semantic,gu2023learnability,wang2023towards}. They show promising results but require extra efforts from stakeholders like LLM providers, not applicable to non-watermarked LLMs. We focus on passive detection, which is more flexible as it operates without modifying LLM workflow. They could be categorized as:

\paragraph{White-box detection methods}
exploit information from probabilities, which reflects the essentials of language modeling.
Earlier works use overall probability~\cite{solaiman2019release}, perplexity~\cite{beresneva2016computer,GPTZero}, or entropy~\cite{lavergne2008detecting} of the given text on LMs as features.
Subsequent works focus on finer-grained token-level probabilities~\cite{gehrmann2019gltr,verma2023ghostbuster}. For example, Sniffer~\cite{sniffer} and SeqXGPT~\cite{wang2023seqxgpt} utilize the probability lists of token sequences on each candidate LLM for multiclass detection.
To obtain richer probability information, recent works also consider perturbation of given text on candidate LLMs through mask-filling~\cite{detectgpt} or re-generating~\cite{yang2023dna}.
White-box detection methods generally perform better and more robustly than black-box ones~\cite{wang2023seqxgpt}, but the required access to LLMs' internal states largely limits their application to black-box LLMs. Some variants use other models as proxies but performances drop significantly~\cite{detectgpt}.

\paragraph{Black-box detection methods}
typically mine effective features from the given text based on semantic representation~\cite{hc3,chen-etal-2023-token,zhan2023g3detector} or stylistic expert knowledge~\cite{frohling2021feature,aich2022demystifying}.
Recent works~\cite{yang2023dna,yu2023gpt} compute the distance between the given text and re-generated texts to reflect the familiarity the candidate LLM has with the given text.
Black-box methods have better applicability, but their performance and generalizability generally fall behind white-box ones, especially for multiclass tasks~\cite{li2023deepfake}. Our POGER combines both their advantages of applicability under the black-box setting and effectiveness brought by (estimated) white-box features.

\section{Preliminary Study on Re-Sampling-Based Black-Box AIGT Detection}
\label{sec:ana}
A recent study reveals that multiple re-sampling could be used to infer the internal decoding strategy of LLMs~\cite{ippolito2023reverse} under black-box access. Inspired by this, we propose a naive black-box solution that estimates word generation probabilities (proved effective for white-box detection) using multiple re-sampling on the given text to preliminarily validate the feasibility.

\subsection{A Naive Solution}
\label{sec:re-sampling}
We implement the straightforward full-text re-sampling as the naive solution, which samples multiple times at each position of the given text to compute word probabilities on the black-box LLM.
For a given text $\boldsymbol{x}$, to obtain the word $x_i$'s probability $\hat{p}(x_i|x_{<i})$, we instruct the black-box LLM for $N$ times using the following prompt:

\begin{quote}
    \textit{Please continue writing the following text, starting from the next word: $\{x_{<i}\}$}
\end{quote}
For each prompting, we obtain the generated word at position $i$ by restricting the maximum output length. The estimated probability of $x_i$ given $\{x_{<i}\}$ is computed as the frequency of $x_i$ in the output word set $\{o_j\}_{j=1}^N$:
\begin{equation}
    \hat{p}(x_i|x_{<i})=\frac{1}{N}\sum_{j=1}^{N}\mathbb{I}(o_j = x_i)\text{,}
\end{equation}
where $\mathbb{I}(\cdot)$ is the indicator function.
By repeating the above process for each word in $\boldsymbol{x}$, we obtain an estimated probability list of $\boldsymbol{x}$ on this black-box LLM, denoted as $\hat{\boldsymbol{p}} = \left\{\hat{p}(x_1), \hat{p}(x_2|x_1), \cdots, \hat{p}(x_n|x_{<n})\right\}$. With $\hat{\boldsymbol{p}}$ as an alternative input, we can now use white-box methods to validate the feasibility of re-sampling-based detection.

\subsection{Experimental Settings}
\label{sec:ana_exp_set}
\paragraph{Dataset.}
Our experiments are based on a dataset consisting of 10,608 text items from humans and seven popular open-sourced or API-based LLMs in two scenarios, covering real-world threats such as low-quality content production, news faking, and student cheating.
\tablename~\ref{tab:dataset_stat} details the statistics.

\begin{table}[t]
    \small
    \centering
    \setlength{\tabcolsep}{2.2pt}
    \begin{tabular}{clrrr}
        \toprule
        Domain & Source & \multicolumn{1}{c}{\#Human} & \multicolumn{1}{c}{\#Generated} & \multicolumn{1}{c}{Avg. Words} \\
        \midrule
        \multirow{2}{*}{QA} & Quora & 437 & 3,059 & 151.32 \\
        & Reddit ELI5 & 383 & 2,681 & 162.39 \\
        \midrule
        \multirow{2}{*}{Writing} & IELTS Essay & 218 & 1,526 & 232.28 \\
        & BBC News & 288 & 2,016 & 196.62 \\
        \midrule
        \multicolumn{2}{c}{Total} & 1,326 & 9,282 & 177.67 \\
        \bottomrule
    \end{tabular}
    \caption{Statistics of our AIGT detection datasets.}
    \label{tab:dataset_stat}
\end{table}

We first obtain human-written samples from Quora and ELI5 dataset~\cite{fan2019eli5} for the QA domain and IELTS essay and BBC news dataset for the writing domain~\cite{greene2006practical}, respectively (500 each source). Subsequently, we prompt the seven LLMs with questions or writing instructions from the original datasets, including GPT-2 XL~\cite{gpt2}, GPT-J~\cite{gptj}, LLaMA-2 13B~\cite{llama2}, Alpaca 7B~\cite{alpaca}, Vicuna 13B~\cite{vicuna}, GPT-3.5 Turbo~\cite{chatgpt}, and GPT-4 Turbo~\cite{gpt4} and collects their responses. We group human and AI texts with the same prompts and remove the groups that contain answers expressing rejection or exceeding 350 words. All samples are split into the train/validation/test sets with a 7:2:1 ratio at the group level.

\paragraph{Metrics.}
We compute F1 for each class and Macro F1 (MacF1) to evaluate overall performance.

\begin{figure}[t]
\centering
    \includegraphics[width=\linewidth,trim=5 5 5 5,clip]{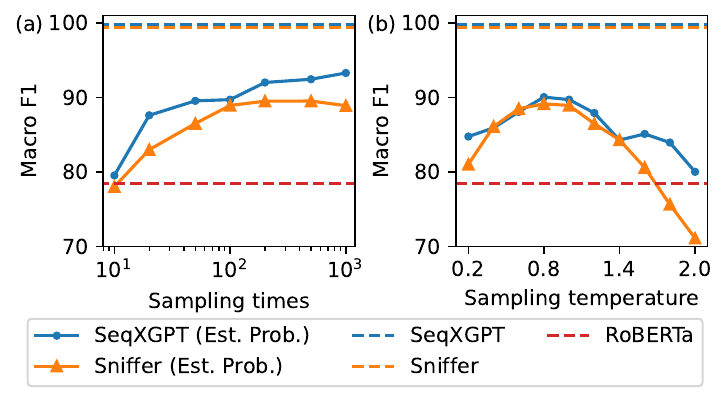}
    \caption{Detection performance using estimated probabilities under different (a) sampling times and (b) sampling temperatures.}
    \label{fig:ana_samples_temp}
\end{figure}

\begin{figure*}[t]
\centering
    \includegraphics[width=\textwidth]{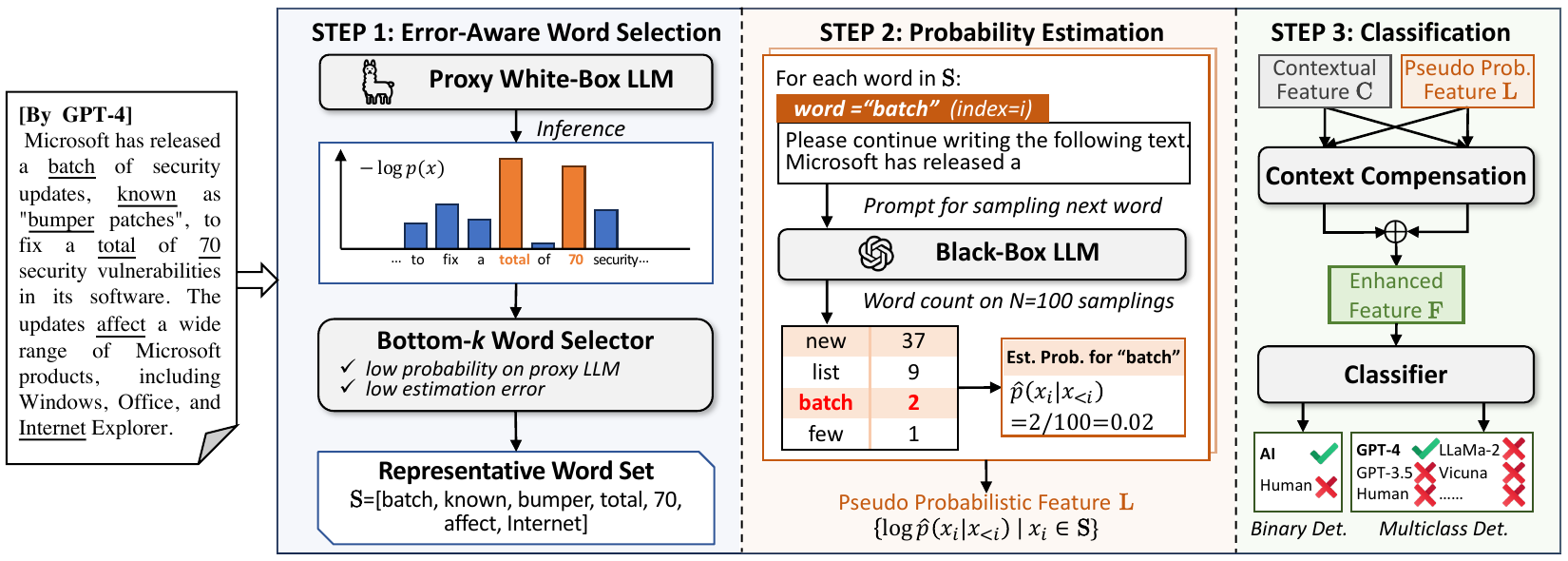}
    \caption{Architecture of \textbf{POGER}. Given a text, POGER operates with three steps: \textbf{1) Error-aware word selection}, where a white-box LLM as a proxy to nominate candidate low-probability words and the bottom-$k$ word selector preserves the lowest $k$ word the satisfied estimation error bound; \textbf{2) Probability estimation}, where multiple re-sampling is applied to candidate black-box LLMs for the selected $k$ word and a pseudo probabilistic feature $\mathbf{L}$ consisting of estimated probabilities is computed; \textbf{3) Classification}, where contextual feature $\mathbf{C}$ is introduced to compensate the context loss in $\mathbf{L}$ to obtain enhanced feature $\mathbf{F}$ for final binary or multiclass AI-generated text detection.}
    \label{fig:method}
\end{figure*}

\subsection{Results \& Analysis}
We implement our naive solution combined with two powerful white-box detectors, Sniffer~\cite{sniffer} and SeqXGPT~\cite{wang2023seqxgpt}. To facilitate comparison between white-box and black-box settings, we only consider the five open-sourced LLMs to easily obtain the true probabilities and set a 6-class task in this part.

\paragraph{Does re-sampling work?} As presented in \figurename~\ref{fig:ana_samples_temp}(a), detection performance increases as the sampling times increase for both Sniffer and SeqXGPT. At sampling times of 100, the macro F1 exceeds that of black-box baseline RoBERTa by 14.3\%. This result demonstrates the feasibility of estimating word probabilities for black-box AIGT detection, though the requirement of sampling times makes this solution costly.

\paragraph{How does estimation error impact detection performance?}
Inevitably, estimation errors exist with re-sampling of limited times. Though having promising results in \figurename~\ref{fig:ana_samples_temp}(a), the detectors still underperform those using true probabilities. To analyze the impact of estimated errors, we adjust the sampling temperature, which changes the probability differences between words and indirectly influences the error. 
In the resulting Figure~\ref{fig:ana_samples_temp}(b), errors lead to low F1 scores on both left and right sides. On the left, a lower temperature indicates the situation that the target word might not be sampled (estimation probability is 0), resulting in a performance drop. On the right, a higher temperature makes the probabilities of all tokens closer. Even slight sampling randomness causes the ranking of probabilities to change and finally distort the unique characteristics of the LLM. This reveals the importance of error control in probability estimation.

Through the preliminary study, we validated the feasibility of estimated probabilities obtained by re-sampling for black-box AIGT detection. We also identify two issues of the naive solution: 1) the sampling cost on the full text is extremely high; 2) the influence of estimation error is not well controlled. Our improved method to be introduced POGER will tackle these issues.

\section{POGER: Proxy-Guided Efficient Re-Sampling for Black-Box AIGT Detection}
To tackle the cost and error control issues exposed in Section~\ref{sec:ana} and design a more practical black-box detector, we propose POGER.
\figurename~\ref{fig:method} presents the overall architecture of POGER. It operates with three steps: First, POGER forms a small word subset from the given text by selecting words of low probabilities and low estimation errors. Subsequently, multiple re-sampling is applied to words in the subset only to obtain a pseudo probabilistic feature. Finally, the feature is enhanced by compensating contextual information and then fed into a classifier for final detection. Details are as follows.

\subsection{Error-Aware Word Selection}
\paragraph{Proxy-Based Candidate Nomination. }
To lower sampling times, we aim to select a small subset of $k$ words from $\boldsymbol{x}$ that reflect the LLM's unique word use characteristics.
Here, we use an easy-to-use LM (\emph{e.g.}, GPT-2) as the proxy for candidate nomination.
The intuition is as follows: As LLMs are usually trained on large-scale human language corpora, they would learn well on common word use and even different LLMs may output similar texts with high probabilities; in contrast, other words with a lower probability in a text are more likely to expose the unique word use shaped by nuances of LLM training process.
Specifically, we use a proxy LM $\theta$ to infer on the given text $\boldsymbol{x}$ and obtain token probabilities on it. We transform the list into word-level by computing joint probabilities of corresponding tokens for multi-token words, denoted as $\boldsymbol{p}^\theta=(p_{1}^\theta,p_{2}^\theta,\cdots,p_{n}^\theta)$. A lower $p_i^\theta$ would make the $x_i$ more likely to be selected.

\paragraph{Error-aware bottom-k Word Selection.}
To mitigate the negative impacts of estimated errors on feature effectiveness, we adopt an error-aware bottom-$k$ word selector.
For a word $x_i$ with true probability $p_i$, if the estimated probability of $x_i$ obtained by re-sampling $N$ times is $\hat{p}_i$, the standard error (SE) of $\hat{p}_i$ is given by:
\begin{equation}
    \mathrm{SE}(\hat{p}_i)=\sqrt{\frac{p_i(1-p_i)}{N}}\text{.}
\end{equation}

For low-probability words, we constrain a lower bound on their true probability to ensure that the error in the estimated probability does not exceed $\Delta$ times itself:
\begin{equation}
    \mathrm{SE}(\hat{p}_i)\leq\Delta\cdot p_i \quad\Rightarrow\quad p_i\geq\frac{1}{1+N\Delta^2}\text{.}
\end{equation}

By controlling the relative error, we remove the items in $\boldsymbol{p}^\theta$ that do not meet the error requirements and obtain $\boldsymbol{p}^{\theta\prime}$. Words which are with lowest $k$ probabilities are selected from $\boldsymbol{p}^{\theta\prime}$ using $\mathrm{MINK}(\cdot)$ function and finally form the representative word set $\mathrm{S}$:
\begin{equation}
    \boldsymbol{p}^{\theta\prime}=\left\{p_i\middle| p_i\geq\frac{1}{1+N\Delta^2}\right\}\text{,}
\end{equation}
\begin{equation}
    \mathrm{IDX} = \left\{i\middle| p_i^\theta \in\mathrm{MINK}(\boldsymbol{p}^{\theta\prime})\right\}\text{,} \ 
    \mathrm{S}=\left\{x_i\middle| i\in \mathrm{IDX}\right\}\text{.}
\end{equation}

\begin{table*}[t]
    \centering
    \small
    \setlength{\tabcolsep}{6.5pt}
    \setlength{\aboverulesep}{0pt}
    \setlength{\belowrulesep}{0pt}
    \newcommand\astrut{\rule[8pt]{0pt}{2pt}}
    \newcommand\bstrut{\rule[-5pt]{0pt}{5pt}}
    \begin{tabular}{lccccccccl}
    \toprule
      \astrut\textbf{Method} & \textbf{Human} & \textbf{GPT-2} & \textbf{GPT-J} & \textbf{LLaMA-2} & \textbf{Vicuna} & \textbf{Alpaca} & \textbf{GPT-3.5} & \textbf{GPT-4} & \multicolumn{1}{c}{\textbf{MacF1}\quad}\bstrut \\
    \midrule
    \midrule
        \multicolumn{10}{c}{\astrut\textbf{Partial White-Box Setting}}\bstrut \\
        \midrule
         \astrut DNA-GPT White & N/A & 62.70 & 40.79 & 45.36 & 30.49 & 70.18 & \cellcolor[gray]{0.9}N/A & \cellcolor[gray]{0.9}N/A & \ 49.91*  \\
         Sniffer & 96.60 & \textbf{100.00} & \textbf{100.00} & \underline{98.49} & 95.85 & \textbf{99.23} & \cellcolor[gray]{0.9}75.34 & \cellcolor[gray]{0.9}72.65 & \ 92.27  \\
        SeqXGPT & \textbf{98.07} & \textbf{100.00} & \underline{99.62} & \textbf{98.88} & \textbf{99.62} & \underline{98.87} & \cellcolor[gray]{0.9}85.93 & \cellcolor[gray]{0.9}84.17 & \ 95.64  \\
        \hdashline
        POGER-Mixture & \underline{97.32} & 98.88 & 99.23 & 98.11 & \underline{97.71} & 98.86 & \cellcolor[gray]{0.9}\textbf{97.36} & \cellcolor[gray]{0.9}\textbf{97.38} & \ \textbf{98.11}   \\
        \quad \textit{w/o CC} &  96.97 & \underline{99.62} & 99.23 & 96.68 & 94.94 & 98.48 & \cellcolor[gray]{0.9}\underline{95.42} & \cellcolor[gray]{0.9}\underline{95.13} & \ \underline{97.06} \bstrut \\
        \midrule
        \multicolumn{10}{c}{\astrut\textbf{Black-Box Setting}}\bstrut \\
        \midrule
        \astrut RoBERTa & 88.24 & \cellcolor[gray]{0.9}78.03 & \cellcolor[gray]{0.9}86.55 & \cellcolor[gray]{0.9}55.47 & \cellcolor[gray]{0.9}58.70 & \cellcolor[gray]{0.9}59.91 & \cellcolor[gray]{0.9}70.63 & \cellcolor[gray]{0.9}84.13 & \ 72.71  \\
        T5-Sentinel & 87.29 & \cellcolor[gray]{0.9}85.42 & \cellcolor[gray]{0.9}\underline{88.71} & \cellcolor[gray]{0.9}67.78 & \cellcolor[gray]{0.9}62.11 & \cellcolor[gray]{0.9}69.73 & \cellcolor[gray]{0.9}75.79 & \cellcolor[gray]{0.9}79.83 & \ 77.08  \\
        DNA-GPT Black & N/A & \cellcolor[gray]{0.9}38.58 & \cellcolor[gray]{0.9}21.56 & \cellcolor[gray]{0.9}48.80 & \cellcolor[gray]{0.9}33.85 & \cellcolor[gray]{0.9}47.15 & \cellcolor[gray]{0.9}53.99 & \cellcolor[gray]{0.9}39.82 & \ 40.53* \\
        Sniffer & 87.41 & \cellcolor[gray]{0.9}\underline{89.82} & \cellcolor[gray]{0.9}87.26 & \cellcolor[gray]{0.9}29.52 & \cellcolor[gray]{0.9}47.62 & \cellcolor[gray]{0.9}35.84 & \cellcolor[gray]{0.9}34.21 & \cellcolor[gray]{0.9}52.63 & \ 58.04 \\
        SeqXGPT & \underline{91.67} & \cellcolor[gray]{0.9}89.66 & \cellcolor[gray]{0.9}86.77 & \cellcolor[gray]{0.9}23.64 & \cellcolor[gray]{0.9}46.31 & \cellcolor[gray]{0.9}45.64 & \cellcolor[gray]{0.9}42.10 & \cellcolor[gray]{0.9}62.40 & \ 61.02 \\
        \hdashline
        POGER & \textbf{92.49} & \cellcolor[gray]{0.9}\textbf{93.75} & \cellcolor[gray]{0.9}\textbf{89.96} & \cellcolor[gray]{0.9}\textbf{90.49} & \cellcolor[gray]{0.9}\textbf{89.30} & \cellcolor[gray]{0.9}\textbf{93.82} & \cellcolor[gray]{0.9}\textbf{90.98} & \cellcolor[gray]{0.9}\textbf{92.59} & \ \textbf{91.67}   \\
        \quad \textit{w/o CC} & 84.21 & \cellcolor[gray]{0.9}88.30 & \cellcolor[gray]{0.9}80.63 & \cellcolor[gray]{0.9}\underline{81.88} & \cellcolor[gray]{0.9}\underline{88.65} & \cellcolor[gray]{0.9}\underline{91.95} & \cellcolor[gray]{0.9}\underline{89.49} & \cellcolor[gray]{0.9}\underline{87.35} & \ \underline{86.56} \bstrut \\
     \bottomrule
    \end{tabular}
    \caption{F1 scores in two settings for multiclass AIGT detection. The best two results are respectively \textbf{bolded} and \underline{underlined}. The \colorbox[gray]{0.9}{shaded area} denotes the performance on black-box LLMs. * Because of the nature of DNA-GPT, the macro F1 scores of DNA-GPT White and Black are derived under a pure-white-box setting (5 classes) and a black-box setting without human class (7 classes). CC: Context Compensation.}
    \label{tab:main_result}
\end{table*}

\subsection{Probability Estimation}
We again use the sampling and probability calculation process described in Section~\ref{sec:re-sampling}, but only for the selected $k$ words in $\mathrm{S}$ on the given $M$ candidate black-box LLMs (denoted as $\{\theta_i\}_{i=1}^M$) by $N$ times.
For efficiency needs, we constrain the maximum context length as $b$. We get the pseudo log probabilistic feature matrix $\mathbf{L}=[\boldsymbol{l}_i]_{i=1}^{k}  \in \mathbb{R}^{k\times M}$, where the $M$-dimentional feature vector for the $i$-th word is $\boldsymbol{l}_i=\left[\hat{p}_{\theta_j}\left(x_{\mathrm{IDX}[i]}|x_{\mathrm{IDX}[i]-b:\mathrm{IDX}[i]-1}\right)\right]_{j=1}^M$.

\subsection{Context-Compensated Classification}
In the final step, the source of the target text $\boldsymbol{x}$ is classified based on $\mathbf{L}$. Following~\cite{wang2023seqxgpt}, we first transform the $\mathbf{L}$ into another $\mathbf{L}' \in \mathbb{R}^{k\times d}$ using convolutional neural network and Transformer to enrich the representation.
Furthermore, since we discontinuously select representative words, the information about the local context around the word and their relative positions in the original text is lost. As context compensation, we introduce the contextual semantic representation of the $k$ words to bootstrap the probabilistic representation $\mathbf{L}'$.

Specifically, we input the given text $\boldsymbol{x}$ into RoBERTa~\cite{roberta} and obtain the last-layer word representation as $\mathbf{E}=\left\{\mathbf{e}_1, \mathbf{e}_2, \cdots, \mathbf{e}_n\right\}$.
The representations for the $k$ representative words are then mapped to $d$ dimension through a multi-layer perception (MLP), forming the contextual feature matrix $\mathbf{C}\in \mathbb{R}^{k\times d}$:
\begin{equation}
    \mathbf{C}=\left[\mathrm{MLP}(\mathbf{e}_i)\right]_{i\in\mathrm{IDX}}\text{.}
\end{equation}
Then a bidirectional cross-attention is adopted to build interaction between $\mathbf{C}$ and $\mathbf{L}'$, outputting the enhance feature $\mathbf{F}\in\mathbb{R}^{k\times 2d}$:
\begin{equation}
    \mathrm{Att}(\mathbf{Q}, \mathbf{K}, \mathbf{V}) = \mathrm{softmax}\left(\frac{\mathbf{Q}\mathbf{K}^T}{\sqrt{d}}\right)\mathbf{V}\text{,}
\end{equation}
\begin{equation}
\mathbf{F}=\mathrm{Att}\left(\mathbf{L}', \mathbf{C}, \mathbf{C}\right)\oplus\mathrm{Att}\left(\mathbf{C}, \mathbf{L}', \mathbf{L}'\right)\text{,}
\end{equation}
where $\oplus$ is a concatenation operation.
Since the representation of each position in $\mathbf{C}$ implies contextual information, this interaction allows relative positional information to be fused into the final enhanced feature. 

Finally, $\mathbf{F}$ is fed into another MLP for final classification. The network is optimized using cross-entropy loss.

\section{Evaluation}
\label{sec:evaluation}

\subsection{Experimental Settings}
\paragraph{Settings.} We continue using the datasets and LLMs introduced in Section~\ref{sec:ana}. To simulate real-world situations, we have two settings: 1) \textbf{Partial White-Box Setting} provides true probabilities for the five open-sourced LLMs; and 2) \textbf{Black-box Setting} treats all models as black-box LLMs.
For the former setting, we provide a variant \textbf{POGER-Mixture}, which uses true probabilities from white-box LLMs and estimated ones from black-box LLMs. The estimated probability list is expanded to the same dimension as the true lists by padding with 0. We evaluate for both binary and multiclass detection tasks.

\paragraph{Baselines.}
\textbf{1) GPTZero}~\cite{GPTZero}: Distinguish between human and generated text using perplexity and burstiness of text;
\textbf{2) RoBERTa}~\cite{hc3}: A widely used and powerful pre-training-based detector;
\textbf{3) T5-Sentinel}~\cite{chen-etal-2023-token} Another pretraining-based method for reframing the classification task as a next-token prediction task.
\textbf{4) DNA-GPT}~\cite{yang2023dna}: Determine the source of text based on multiple re-generation, can works in two forms under black-box and white-box settings.
\textbf{5) DetectGPT}~\cite{detectgpt}: Determine whether a text is generated by comparing the probability of the original text with a large number of perturbed texts.
\textbf{6) Sniffer}~\cite{sniffer}: Determine the source of the text using the contrastive features of the probability lists on each candidate model.
\textbf{7) SeqXGPT}~\cite{wang2023seqxgpt}: Also based on probability lists of the text, but reframes the classification task as a sequence labeling task.

\paragraph{Implementation Details.}
For POGER and POGER-Mixture, we use GPT-2 Large as the proxy for representative word selection, with maximum error tolerance $\Delta=1.2$, representative word set size $k=10$, re-sampling times $N=100$, and sampling temperature $t=1.0$.
For the five open-source models, we perform re-sampling locally, and for GPT-3.5 Turbo and GPT-4 Turbo, we call OpenAI API for re-sampling and set \textit{max\_tokens} to 2. We input $b=20$ words before the target word as context.
For white-box detection methods under the black-box setting, we employ GPT-Neo 2.7B and LLaMA 7B as proxy probability providers to ensure their proper functionality (some methods require at least two proxies).
For zero-shot detection methods, we do a grid search on classification thresholds and report their optimal performance in the search interval.

\subsection{Main Results}
\subsubsection{Multiclass AIGT Detection}
Table~\ref{tab:main_result} shows the performance comparison of POGER and its variants with other baselines. Based on the result, we have the following observations:

\begin{compactitem}
     \item POGER and POGER-Mixture outperform all baselines in macro F1 in both the black-box and partial white-box settings. In particular, POGER and POGER-Mixture outperform all black-box LLMs (shaded in the table) in single-class F1. This demonstrates that POGER has superior performance, especially for black-box detection.

    \item Compared with the partial white-box setting, all baseline models experience significant performance degradation in the black-box setting. Among the baseline models, semantic-based RoBERTa and T5-Sentinel perform the best, but they still fall behind POGER by more than 15.9\% in macro F1, which indicates that POGER could achieve a good balance between applicability and performance.

    \item  We evaluate the effectiveness of the Context Compensation module in POGER and POGER-Mixture in both settings. We can see that, on the one hand, \textit{w/o Context Compensation} 
    brings a decrease of over 5 macro F1 scores in POGER performance, highlighting the significance of this module.
    On the other hand, even the \textit{w/o Context Compensation} variant of POGER still outperforms all baselines, demonstrating the effectiveness of the resampling strategy we proposed.
\end{compactitem}

\subsubsection{Binary AIGT Detection}
\begin{table}[t]
    \centering
    \small
    \setlength{\tabcolsep}{2.4pt}
    \begin{tabular}{@{}clccc@{}}
        \toprule
        \textbf{Setting} & \textbf{Method} & \textbf{Human} & \textbf{Generated} & \textbf{MacF1} \\
        \midrule
        \multirow{4}{*}[0pt]{\makecell{Partial\\White-Box}} & DetectGPT & 60.00 & 95.58 & 77.79 \\
        & DNA-GPT White & 77.05 & 97.00 & 87.03 \\
        & SeqXGPT & 96.60 & 99.51 & 98.06 \\
        & POGER-Mixture & \textbf{97.69} & \textbf{99.67} & \textbf{98.68} \\
        \midrule
        \multirow{7}{*}[0pt]{Black-Box} & RoBERTa & 92.06 & 98.92 & 95.49 \\
        & T5-Sentinel & 87.29 & 98.40 & 92.85 \\
        & DNA-GPT Black & 42.08 & 87.09 & 64.59 \\
        & DetectGPT & 44.81 & 92.89 & 68.85 \\
        & SeqXGPT & 92.07 & 98.86 & 95.47 \\
        & GPTZero & 68.42 & 95.45 & 81.94 \\
        & POGER & \textbf{93.89} & \textbf{99.14} & \textbf{96.51} \\
        \bottomrule
    \end{tabular}
    \caption{F1 scores in two settings of binary AIGT detection. The best result under each setting is \textbf{bolded}.}
    \label{tab:binary_result}
\end{table}

Table~\ref{tab:binary_result} shows the performance of each method in binary detection. Among the baselines, RoBERTa, SeqXGPT, and T5-Sentinel
perform well (macro F1 of over 90) while others gain unsatisfying performance. This might be influenced by the method nature of focusing more on distinguishing different LLMs. Still, POGER gains the best performance in both settings, showing its wide applicability in different AIGT detection tasks.

\subsection{Out-of-Distribution Results}
Due to variations in semantic and stylistic features across different domains in AIGT, existing training-based black-box detection methods often exhibit poor performance in Out-Of-Distribution (OOD) scenarios. We conducted multi-class AIGT detection experiments between two domains within our dataset, where training samples were sourced from one domain and testing was performed on the other domain. The results are presented in Table~\ref{tab:ood}.

\begin{table}[t]
    \centering
    \small
    \setlength{\tabcolsep}{2.8pt}
    \begin{tabular}{lccrcr}
        \toprule
        \multirow{2}{*}[-0.3em]{\textbf{Method}} & \multirow{2}{*}[-0.3em]{\textbf{In-Dist.}} & \multicolumn{4}{c}{\textbf{Out-of-Distribution}} \\
        \cmidrule{3-6}
        & & \multicolumn{2}{c}{QA$\rightarrow$Writing} & \multicolumn{2}{c}{Writing$\rightarrow$QA} \\
        \midrule
        RoBERTa & 72.71 & 54.23 & \textit{(-25.42\%)} & 46.73 & \textit{(-35.73\%)} \\
        T5-Sentinel & 77.08 & 47.23 & \textit{(-38.73\%)} & 53.19 & \textit{(-30.99\%)} \\
        Sniffer & 58.04 & 57.50 & \textit{(-0.93\%)} & 53.16 & \textit{(-8.41\%)} \\
        SeqXGPT & 61.02 & 59.07 & \textit{(-3.20\%)} & 54.94 & \textit{(-9.96\%)} \\
        POGER & \textbf{91.67} & \textbf{89.00} & \textit{(-2.91\%)} & \textbf{84.19} & \textit{(-8.16\%)} \\
        \bottomrule
    \end{tabular}%
    \caption{F1 scores of the OOD experiment. The relative decrease for OOD scenarios over the in-distribution F1 score is shown in the brackets. In-Dist.: In-Distribution.}
    \label{tab:ood}
\end{table}

It can be seen that POGER still outperforms all baselines, both in terms of generalizing from QA to writing and vice versa. Meanwhile, compared with their respective In-Distribution performance, the relative performance degradation of POGER in the OOD scenario is significantly smaller than the two black-box baselines RoBERTa and T5-Sentinel, and is comparable to the two white-box baselines Sniffer and SeqXGPT.
This indicates that, despite being a black-box AIGT detector, POGER benefits from the excellent OOD generalization capabilities inherited from white-box detection methods through the pseudo probabilistic feature.

\section{Analysis}
\begin{figure}[t]
    \centering
    \begin{minipage}{0.48\linewidth}
        \centering
        \includegraphics[width=\linewidth,trim=10 10 0 0,clip]{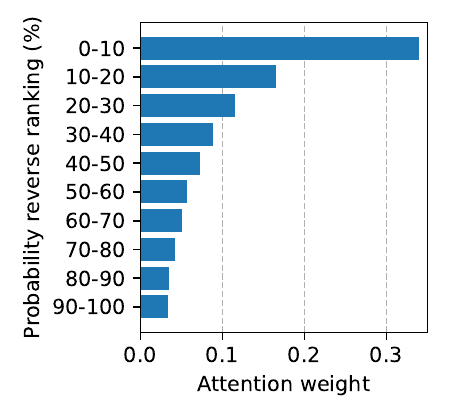}
        \caption{Distribution of attention weight for words in different probability ranking intervals.}
        \label{fig:attention}
    \end{minipage}%
    \hspace{0.02\linewidth}
    \begin{minipage}{0.48\linewidth}
        \centering
        \includegraphics[width=\linewidth,trim=10 10 0 0,clip]{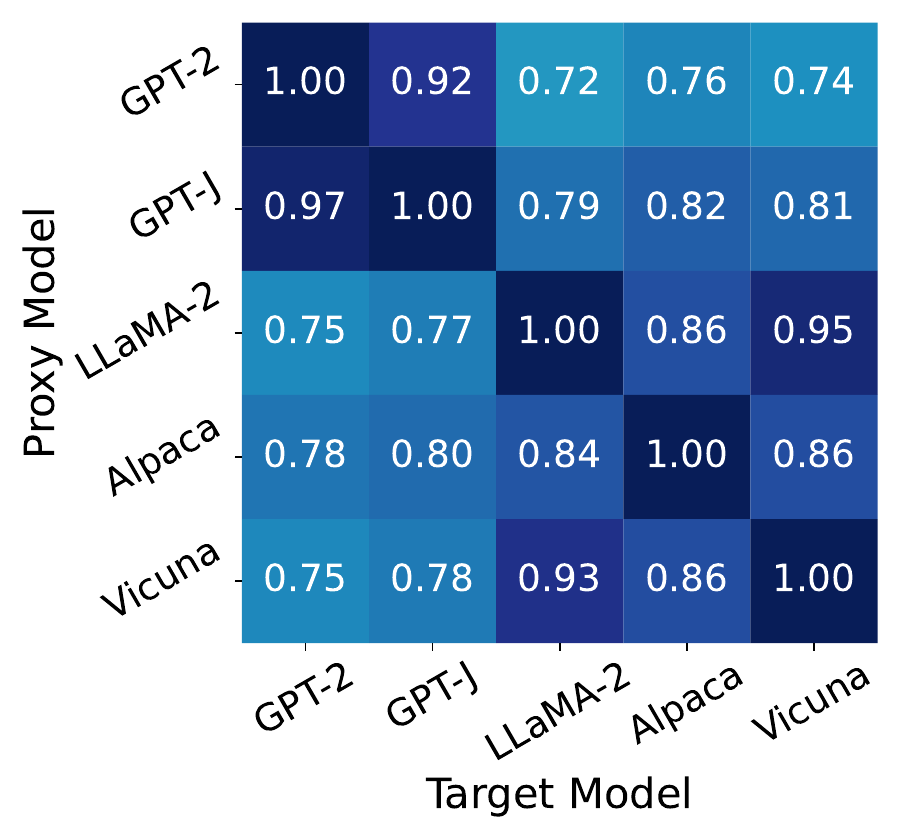}
        \caption{Overlapping Proportion of low-probability words between different LLMs.}
        \label{fig:heatmap}
    \end{minipage}
\end{figure}

\subsection{Representativeness of Selected Words}
We conduct empirical analysis to validate if our word selection methods will select representative words.

\paragraph{Are low-probability words more helpful in detection?} We implement a simple white-box AIGT detector with an attention layer that takes true probabilities on candidate LLMs as inputs. Figure~\ref{fig:attention} shows the attention distribution on words in different probability ranking intervals. We observe that the 10\% lowest-probability words gain over 30\% of the attention weights. As word probability increases, their attention weights decline, indicating low-probability words play more important roles in AIGT detection.

\paragraph{Are low-probability word sets similar between the proxy and candidate LLMs?} We use each LLM as a proxy model and the other LLMs as target models and show the proportion of words with the lowest 5\% probability on the proxy model whose probability on the target model was in the lowest 20\% in Figure~\ref{fig:heatmap}. Even in the worst case, 72\% of low-probability words on the proxy model are also hit in the set of the target model. This suggests that proxy model could be used as a good indicator for word selection.

\subsection{Hyperparameter Senstivity}
\begin{figure}[t]
\centering
    \includegraphics[width=\linewidth,trim=5 5 5 5,clip]{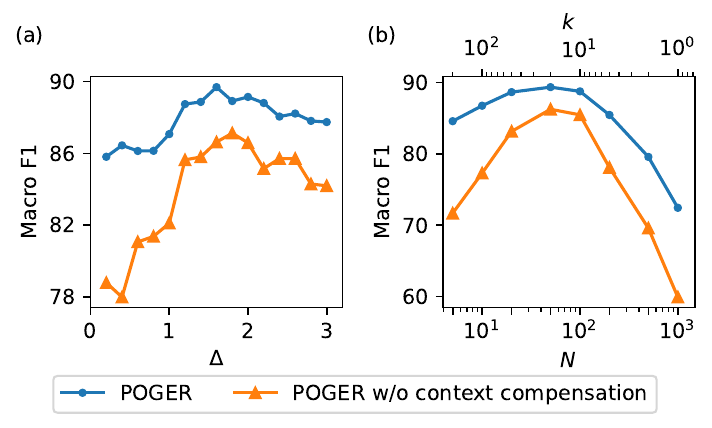}
    \caption{Performance of POGER and its variant under (a) different maximum error tolerances (b) different representative word set sizes \& re-sampling times, where the product of $N$ and $k$ is constrained to be equal to 1,000.}
    \label{fig:hyperparams}
\end{figure}

We conduct an analysis of four hyper-parameters on a subset containing text from humans and the five open-sourced LLMs for brevity:
\paragraph{Impact of Maximum Error Tolerance.}
Figure~\ref{fig:hyperparams}(a) shows the performance impact of the maximum error tolerance $\Delta$ in the representative word selector. As $\Delta$ increases, we find that the performance curve shows a trend of rising first and then falling. This is due to the fact that when $\Delta$ takes a small value, as the selector has a strict constraint on the error of empirical probability, the low-probability words that best reflect the characteristics of the text source are filtered out, rendering the words in set $\mathrm{S}$ insufficiently representative. Representative; when $\Delta$ takes a large value, the error of empirical probability also increases, resulting in the weakening of the effectiveness of probabilistic features. When $\Delta$ is between 1.2 and 2.2, our selector can strike a good balance between word representativeness and estimation error.

\paragraph{Impact of Representative Word Set Size \& Re-Sampling Times.}
Intuitively, if the representative word set size $k$ and the re-sampling time $N$ are increased at any cost, the performance of the detector will also increase. However, in practice, we expect POGER to achieve the best possible detection performance while meeting certain cost and efficiency requirements. Therefore, we examine the effect of $k$ and $N$ on POGER performance at the same level of total sampling number on the black-box LLM.
Figure~\ref{fig:hyperparams}(b) shows the performance of POGER under different $(k, N)$ pairs when the total sampling number $k\cdot N=1000$. It is observed that POGER performs best when $k$ is between 10 and 50 (\emph{i.e.}, $N$ is between 20 and 100). When $k$ is too small, the detector obtains too little information about probability lists; and when $N$ is too small, the error-aware selector filters out a large number of words with excessive errors, resulting in the selected words not being representative enough. Both situations result in POGER failing to make accurate classifications.

We find that compared to the version without content compensation, the full POGER has less variation in performance in response to changes in either $N$, $k$, or $\Delta$. We believe this is due to the fact that the contextual semantic information is able to compensate for a portion of the performance in cases where the probabilistic features are not effective enough.

\paragraph{Proxy Model for Representative Words Selection.}
We conduct small-scale experiments using GPT-2 Large, GPT-Neo 2.7B, and LLaMA 7B as proxy models in representative word selection, and the Macro F1s are 85.77, 85.42, and 86.14, respectively, indicating that POGER is not sensitive to the selection of proxy models. It also proves once again that not using the probability value of the proxy model but determining the approximate range of the probability is a proper way of using proxy models for AIGT detection.

\subsection{Cost Comparison}
With a well-designed re-sampling strategy, we achieve high-performance black-box AIGT detection at a relatively low cost. Although POGER requires re-sampling for each representative word, we only infer for the next position to obtain the generated word instead of repeatedly generating the whole text sequence, thus requiring limited LLM inference cost. If the maximum number of generated tokens is set to 1, the inference length of LLM is even independent of the re-sampling times, since no additional inference beyond prompt is required. Table~\ref{tab:cost} shows a comparison of the number of inference tokens on a target LLM for POGER and other regeneration-based AIGT detection methods, indicating that POGER's inference cost is similar to DNA-GPT and much less than DetectGPT and Full Sampling. 

\begin{table}[t]
    \centering
    \small
    \setlength{\tabcolsep}{2.2pt}
    \begin{tabular}{@{}cccc@{}}
        \toprule
        \multirow{2}{*}[-0.3em]{\textbf{Method}} & \multicolumn{3}{c}{\textbf{\# Target LLM Inference Tokens}} \\
        \cmidrule{2-4}
        & Expression & Typical Value & Ratio \\
        \midrule
        DetectGPT & $n\cdot l$ & 30,000 & $\times$ 21.43 \\
        DNA-GPT & $r\cdot l+n\cdot(1-r)\cdot l$ & 1,650 & $\times$ 1.18 \\
        Full Sampling & $l\cdot \left[l_p+n(m-1)\right]$ & 42,000 & $\times$ 30 \\
        POGER & $k\cdot \left[l_p+n(m-1)\right]$ & 1,400 & $\times$ 1 \\
        \bottomrule
    \end{tabular}%
    \caption{Comparison of the number of inference tokens needed for detection, where $n$ denotes the number of re-generated samples, $l$ denotes the text length, $r$ denotes the truncation ratio in DNA-GPT, $k$, $l_p$, and $m$ denote the size of the representative word set, the prompt length, and the maximum number of generated tokens in POGER, respectively. Full Sampling refers to the naive solution in Section~\ref{sec:re-sampling}. In the calculation of typical values, $l$ is taken as 300 tokens (about 200 words), and the values of other variables are referred to the original publication.}
    \label{tab:cost}
\end{table}

\section{Conclusion and Discussion}
In this paper, we proposed to estimate features that proved effective in the white-box setting to help improve black-box AIGT detection. We first developed a naive solution that leverages multiple re-sampling to estimate word generation probabilities for black-box detection. To further reduce the sampling cost, we designed POGER, which leverages a proxy model to select a subset of representative words with an awareness of sampling errors. Experiments on texts from humans and seven LLMs demonstrated the superiority of POGER for binary, multiclass, and OOD scenarios. Further cost analysis indicates that POGER keeps lower re-sampling costs than its counterparts. In the future, we plan to further improve the efficiency by introducing result storage.

\section*{Ethical Statement}
Considering that white-box and black-box LLMs like LLaMA-2~\cite{llama2} and GPT-4~\cite{gpt4} have been widely used in people's daily lives, and the AI-generated texts have posed real-world threats and are believed to bring more serious societal harms, our research attempts to propose a new method for AI-generated text detection to help defense against the unknown threats in the future. Considering the performance is not perfect for now, the text that is flagged as AI-generated text by POGER and its counterparts should go through extra checking before making an official accusation regarding rule violations to a certain person in practice.

\bibliography{custom}

\appendix

\section{Dataset Construction Details}
\begin{table*}[t]
    \centering
    \small
    \begin{tabular}{ccp{10.2cm}}
        \toprule
        \textbf{Source} & \textbf{Model} & \textbf{Prompt} \\
        \midrule
        \multirow{2}{*}[-18pt]{BBC News} & \makecell{GPT-2 \& GPT-J} & \texttt{\makecell[l]{BBC News:\\---title---\\ \textrm{\textit{\{Title\}}}\\---content---}} \\
        \cmidrule{2-3}
        & Others & \texttt{\makecell[l]{\textrm{\textit{System Prompt:}} You are a BBC news editor. I will give you a news \\title. Please write the content of the news, which should not \\exceed 300 words.\\\textrm{\textit{User Prompt:}} Title: \textrm{\textit{\{Title\}}} }} \\
        \midrule
        \multirow{2}{*}[-18pt]{IELTS Essay} & \makecell{GPT-2 \& GPT-J} & \texttt{\makecell[l]{IELTS essay:\\---question---\\ \textrm{\textit{\{Question\}}}\\---essay---}} \\
        \cmidrule{2-3}
        & Others & \texttt{\makecell[l]{\textrm{\textit{System Prompt:}} You are an IELTS candidate. I will give you the \\question that you need to complete the IELTS essay, which should \\not exceed 300 words.\\\textrm{\textit{User Prompt:}} \textrm{\textit{\{Question\}}} }} \\
        \midrule
        \multirow{2}{*}[-3pt]{\makecell{Quora \\ \& \\ Reddit ELI5}} & \makecell{GPT-2 \& GPT-J} & \texttt{\makecell[l]{Question: \textrm{\textit{\{Question\}}}\\Answer:}} \\
        \cmidrule{2-3}
        & Others & \texttt{\makecell[l]{\textrm{\textit{System Prompt:}} Answer the following questions, within 300 words.\\\textrm{\textit{User Prompt:}} \textrm{\textit{\{Question\}}} }} \\
        \bottomrule
    \end{tabular}
    \caption{Prompts used for the AIGT dataset construction.}
    \label{tab:prompt}
\end{table*}

\begin{table*}[ht]
    \centering
    \small
    \setlength{\tabcolsep}{7pt}
    \begin{tabular}{lccccccccl}
    \toprule
      \textbf{Method} & \textbf{Human} & \textbf{GPT-2} & \textbf{GPT-J} & \textbf{LLaMA-2} & \textbf{Vicuna} & \textbf{Alpaca} & \textbf{GPT-3.5} & \textbf{GPT-4} & \multicolumn{1}{c}{\textbf{MacF1}\quad} \\
    \midrule
    \midrule
        \multicolumn{10}{c}{\textbf{QA$\rightarrow$Writing}} \\
        \midrule
        RoBERTa & 65.79 & 70.59 & 80.43 & 21.54 & 50.96 & 69.23 & 50.72 & 24.62 & 54.23 \\
        T5-Sentinel & 59.46 & 63.56 & 69.23 & 36.37 & 38.65 & 54.08 & 30.23 & 26.23 & 47.23 \\
        Sniffer & 72.34 & 90.39 & 90.53 & 42.11 & 42.69 & 41.74 & 27.85 & 52.38 & 57.50 \\
        SeqXGPT & 83.81 & 88.89 & \textbf{92.31} & 17.14 & 41.22 & 42.11 & 42.31 & 64.81 & 59.07 \\
        POGER & \textbf{87.28} & \textbf{94.00} & 84.53 & \textbf{92.93} & \textbf{88.17} & \textbf{93.33} & \textbf{87.38} & \textbf{84.41} & \textbf{89.00} \\
        \midrule
        \multicolumn{10}{c}{\textbf{Writing$\rightarrow$QA}} \\
        \midrule
        RoBERTa & 68.29 & 58.33 & 35.85 & 29.01 & 47.81 & 38.93 & 40.74 & 54.88 & 46.73 \\
        T5-Sentinel & 65.04 & 60.25 & 72.00 & 37.08 & 33.34 & 56.46 & 59.86 & 41.51 & 53.19 \\
        Sniffer & 84.67 & 86.58 & \textbf{81.99} & 25.00 & 45.51 & 27.45 & 36.18 & 37.91 & 53.16 \\
        SeqXGPT & \textbf{94.48} & 82.56 & 79.31 & 13.56 & 46.78 & 34.44 & 35.63 & 52.78 & 54.94 \\
        POGER & 79.52 & \textbf{89.47} & 80.88 & \textbf{88.51} & \textbf{88.63} & \textbf{91.86} & \textbf{78.26} & \textbf{76.39} & \textbf{84.19} \\
     \bottomrule
    \end{tabular}
    \caption{F1 scores for each class under the OOD settings. \textbf{QA$\rightarrow$Writing} (\textbf{Writing$\rightarrow$QA}) denotes training on QA(Writing) data and testing on Writing(QA) data.}
    \label{tab:ood_full_result}
\end{table*} 

We use the prompts presented in Table~\ref{tab:prompt} to collect AI-generated texts (AIGTs) in the dataset, with all LLMs performing generation at a sampling temperature of 0.7.
The specific versions of the OpenAI models 
we used are \textit{gpt-3.5-turbo-0613}\footnote{\url{https://platform.openai.com/docs/models/gpt-3-5}} and \textit{gpt-4-1106-preview}\footnote{\url{https://platform.openai.com/docs/models/gpt-4-and-gpt-4-turbo}}.

\section{Implementation Details}

\subsection{POGER}
Since the tokenizers of each LLM are different, the tokenization results are usually not aligned across different LLMs. To construct a unified model, we use word-level estimated probabilities instead of those at the token level, given that the minimum input unit of a black-box LLM is generally a word.
For the true probability lists used in POGER-Mixture, the token probabilities are aligned to the word probabilities. Specifically, for a word composed of multiple tokens, the word probability is calculated as the joint probability of these tokens. For convenience, all probabilities used in POGER and its variants are transformed into negative logarithm probabilities.

In the context-compensated classification module, for the probabilistic feature extraction, the CNN consists of five convolution kernels with sizes of 5, 3, 3, 3, and 3, respectively, and the transformer network~\cite{vaswani2017attention} follows an encoder-only architecture with 4 attention heads. For contextual semantic feature extraction, we use the RoBERTa-base model~\cite{roberta} implemented in the HuggingFace's \textit{Transformers} package~\cite{huggingface}. The final classification is performed by a multi-layer perceptron (MLP) with two hidden layers. During training, we use an Adam optimizer~\cite{adam} with a batch size of 64 and a learning rate of 1e-4.

\subsection{Compared Methods}
\noindent\textbf{DNA-GPT~\cite{yang2023dna}.}
We set the number of re-generations $N=10$, and the truncation ratio $\gamma=0.5$.
For multiclass AIGT detection, as described in the original publication, we sort the WScore or BScore of the given text on each candidate LLM and select the class with the highest score as the predicted label.
For binary AIGT detection, since it mixes multiple candidate LLMs into an ``AI'' label, we calculate the maximum score of the given text on each candidate LLM, search for the decision threshold $\epsilon$ from 0 to 3 in steps of 1e-5, and report the best performance. Our necessary modification of decision-making makes DNA-GPT applicable to perform binary AIGT detection with multiple AI-generated text sources. Our careful search could expose the almost optimal performance of the compared method and ensure that this would not lead to unfairness for DNA-GPT.

\noindent\textbf{DetectGPT~\cite{detectgpt}.}
We use T5-Large~\cite{2020t5} as the mask-filling model and set the number of perturbations as 100.
For binary AIGT detection, we follow a similar implementation to DNA-GPT. We use the average score on each candidate LLM and search for the decision threshold $\epsilon$ from -0.5 to 2.5 in steps of 1e-5, and report the best performance.

\noindent\textbf{Sniffer~\cite{sniffer} \& SeqXGPT~\cite{wang2023seqxgpt}.}
Under the black-box setting, we employ GPT-Neo 2.7B~\cite{gpt-neo} and LLaMA 7B~\cite{llama} as proxy probability providers to ensure their proper functionality, because they require at least two proxies.

\section{Detailed Results for OOD Experiments}

Table~\ref{tab:ood_full_result} shows the performance of POGER and other training-based baselines in terms of F1 score under the out-of-distribution (OOD) setting. The results show that POGER performs better than the baselines for AIGTs from most candidate LLMs.

\end{document}